\documentclass[a4paper,conference]{IEEEtran}
\usepackage{amsmath,amssymb,amsthm,epsfig,epstopdf,mathrsfs,bm,graphicx,subfigure,caption}
\usepackage[ruled]{algorithm2e}

\newcounter{RomanNumber}

\begin{document}
%
\title{Visual Tree Convolutional Neural Network in Image Classification}


%
\author{\IEEEauthorblockN{Yuntao Liu,
Yong Dou,
Ruochun Jin and
Peng Qiao}
\IEEEauthorblockA{National University of Defense Technology\\
National Laboratory for Parallel and Distributed Processing\\
Hunan, Changsha, China, 410073\\
liuyuntao.me@gmail.com, \{yongdou, jinruochun, pengqiao\}@nudt.edu.cn}}


\maketitle

\begin{abstract}
In image classification, Convolutional Neural Network(CNN) models have achieved high performance with the rapid development in deep learning. However, some categories in the image datasets are more difficult to distinguished than others. Improving the classification accuracy on these confused categories is benefit to the overall performance. In this paper, we build a Confusion Visual Tree(CVT) based on the confused semantic level information to identify the confused categories. With the information provided by the CVT, we can lead the CNN training procedure to pay more attention on these confused categories. Therefore, we propose Visual Tree Convolutional Neural Networks(VT-CNN) based on the original deep CNN embedded with our CVT. We evaluate our VT-CNN model on the benchmark datasets CIFAR-$10$ and CIFAR-$100$. In our experiments, we build up $3$ different VT-CNN models and they obtain improvement over their based CNN models by $1.36\%$, $0.89\%$ and $0.64\%$, respectively.
\end{abstract}


%
\IEEEpeerreviewmaketitle

\section{Introduction}
The CNN models are widely used in image classification tasks and their performance is better than any other traditional methods~\cite{Krizhevsky2012ImageNet}. Though the classification accuracy of the state-of-the-art model is surpassing that of human beings, there still remains a challenge that it is difficult for CNN models to discriminate the categories with high visual similarity. We know that some instances are difficult to distinguished while they come from different categories. The misclassification between instances from these categories makes a great contribution to the remaining error rate of CNN models. Take categories in the ImageNet\cite{Russakovsky2015ImageNet} dataset as an example. We consider a coarse-grained category set ''Dog'' and the fine-grained categories which contain $120$ different dog species such as ''Basenji'', ''Pembroke'' and ''Cardigan'' in this set. Compared with the coarse-grained category sets, the fine-grained categories in each set have strong visual confusion~\cite{Jin2017Confusion}. We can easily distinguish two coarse-grained category sets such as "Dog" and "Chariot", but it is difficult to discriminate the $120$ fine-grained categories in "Dog" set.

There are two reasons for this problem. The one is that the structure of the CNN models may be the limitation of the performance. The structure may be not deep enough to get the higher performance. The other is that the strategy which guides the training of the CNN models is not fit with the fine-grained categories classification in image classification tasks. In this paper, we try to resolve the problem caused by the latter reason. The original CNN models output predictions of all the categories in the dataset at once, which means the models actually treat these categories equally. Therefore, the confused fine-grained categories mentioned above make the distinguishing ability of the original CNN models encounter a bottleneck.

Some traditional methods~\cite{Bengio2010Label}\cite{Fan2015Hierarchical} based on the tree structure are proposed to deal with the problems caused by the confused categories. These traditional methods put coarse-grained categories to low-level layers of the tree structure and fine-grained categories to high-level layers. On the one hand, the benefit is that coarse-grained categories are easily to distinguished so the tree classifiers can finish this classification quickly and effectively. On the other hand, the tree classifiers can mainly focus on different sets of fine-grained categories, which is the most difficult part of the classification tasks.

Inspired by the tree classifier methods, we can also combine CNN models with the tree structure to deal with the problem of imbalance among quantities of image categories. Our idea is to construct new CNN models by embedding a visual tree constructed on categories in the dataset into original CNN models. We utilize the visual tree to guide the training of the CNN models. We want the CNN models mainly focus on the category of one sample and the confused categories of it when doing the fine-grained categories classification. Notice that we can get the confused categories from the fine-grained categories on the high-level layers of the visual tree. Different with the tree classifier, our CNN models should output predictions of all the categories in the dataset instead of a part of that, which means that our models focus on the confused categories of one sample while inhibiting other categories. Therefore, we need the outputs from coarse-grained categories classification of our CNN models. We can achieve it by utilizing the coarse-grained categories on the low-level layers in the visual tree.

\begin{figure*}[t]
\subfigure[]{
\label{fig:figure1a}
\begin{minipage}[t]{0.6\linewidth}
  \centering
  \centerline{\includegraphics[width=12cm]{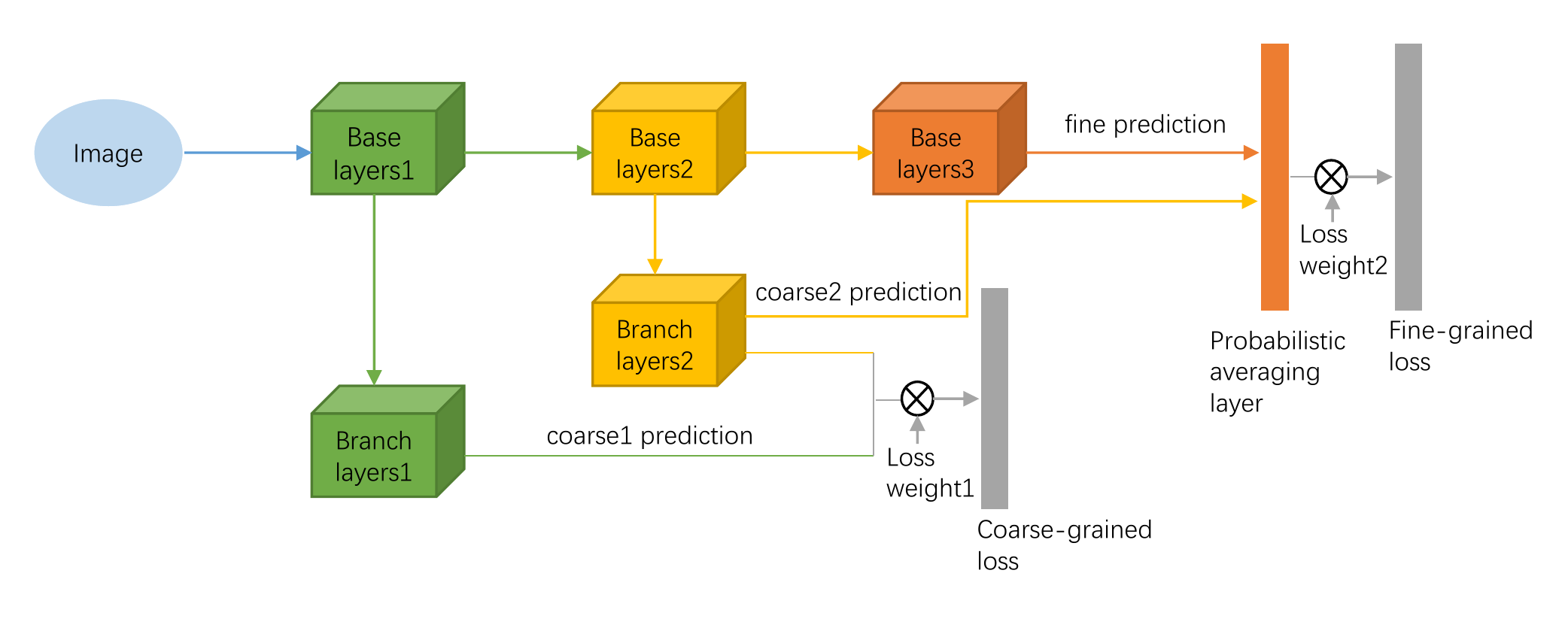}}
\end{minipage}
}
\subfigure[]{
\label{fig:figure1b}
\begin{minipage}[t]{0.4\linewidth}
  \centering
  \centerline{\includegraphics[width=6cm]{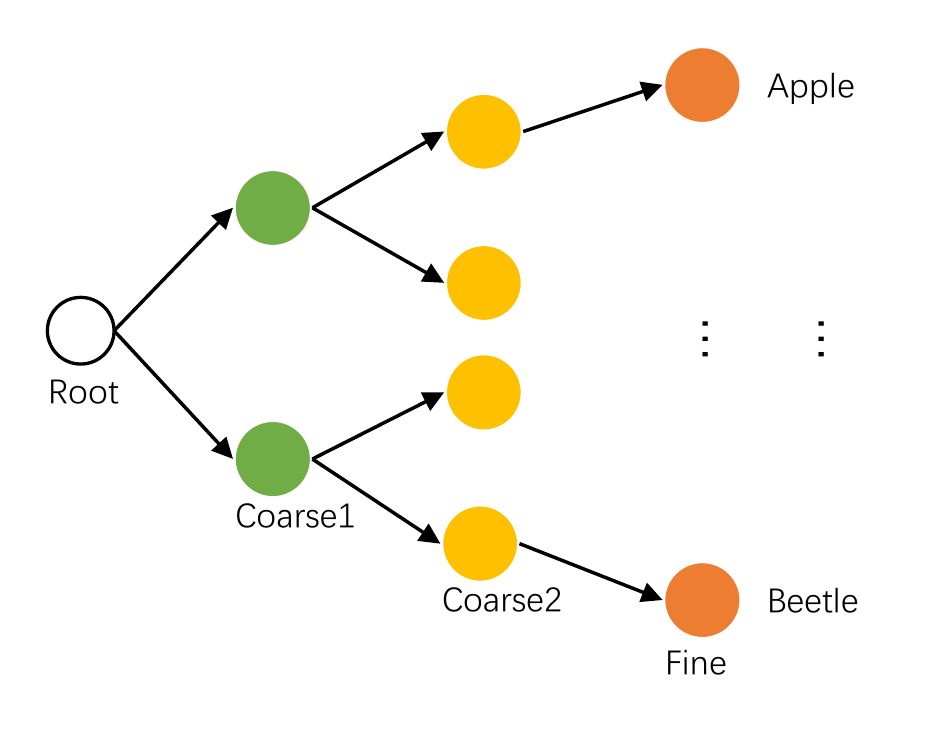}}
\end{minipage}
}
\caption{(a) A four-level Confusion Visual Tree where the classes are taken from CIFAR-$100$ dataset. (b) Visual Tree Convolutional Neural Network(VT-CNN) architecture.}
\label{fig:figure1}
\end{figure*}

Considering the issues mentioned above, we should firstly construct a visual tree structure in order to provide the coarse-grained categories and the fine-grained category seperations for CNN models. Then we should change the structure of the original CNN models aiming to fit with the fine-grained confused categories classification in the image classification tasks. In this paper, we propose a novel scheme that combining a visual tree structure to the CNN model and we name it {\bf Visual Tree Convolutional Neural Network} ({\bf VT-CNN}). The architecture of the VT-CNN model is shown in Fig. \ref{fig:figure1a}. Constructing the VT-CNN model contains two main steps. The first step is constructing a visual tree for image categories. Inspired by the idea of confusion graph in CNN models from~\cite{Jin2017Confusion}, we propose to use the community hierarchical detection algorithm to construct the visual tree, called {\bf Confusion Visual Tree(CVT)}. With this method, we can construct the visual tree structure automatically and fully utilize the information from the output of the CNN models. Then we embed this tree structure into the CNN model. Borrowing the idea from~\cite{Zhu2017B}, we divide the structure of the VT-CNN model into two parts. The one is the base layers and the other is the branch layers. Branch layers have several branches and each branch contains a series of layers such as convolutional layers and fully-connected(FC) layers . All of these branches share the base layers. The number of the branch layers is equal to the number of levels in the tree model without the root level, which means each branch refers to a layer-grained categories classification. By using the step-by-step training method proposed in~\cite{Simonyan2014Very}, the coarse-grained categories branches are trained earlier than fine-grained categories branches, which can prevent the impact of vanishing gradient problem to an extent and boost the performance of the VT-CNN model.

\section{Related Work}
\label{sec:rw}
A large number of works focus on the tree models construction. Some previous works~\cite{Li2010Building}\cite{Zhao2011Large} have leveraged the semantic ontologies (taxonomies) to organize large number of image categories hierarchically. Some researches~\cite{Bengio2010Label}\cite{Jia2011Fast} learn label trees and probabilistic label trees ~\cite{Liu2013Probabilistic} in order to learn a visual hierarchy directly from large amounts of dataset. While, other researchers have proposed visual tree learning algorithm~\cite{Fan2015Hierarchical}\cite{Fan2012Quantitative} to organize large number of image categories hierarchically and use a hierarchical multi-task sparse metric learning algorithm to learn an enhanced visual tree~\cite{Zheng2017Hierarchical} structure. These methods use various algorithms such as SVM, clustering, confusion matrix and so on. But these algorithms have large computation and their performance crucially depends on the distinguishing ability of the underlying metrics for similarity characterization for the algorithm they used.

Different with these method, our work learns the tree structure directly from the output from CNN models. In~\cite{Jin2017Confusion}, a confusion graph is constructed by using the output of the last FC layer in a deep model. It firstly adds the top $N$ score of each image in a dataset calculated by the last FC layer in a CNN model to generate a complete graph. Then applying community detection algorithm~\cite{Vincent2008Fast} on this graph to generate a confusion graph for this dataset. Inspired by this work, we extend this algorithm to a hierarchical community detection algorithm so we can construct a CVT directly from the output of CNN models.

\begin{figure*}[t]
\begin{minipage}[t]{1.0\linewidth}
  \centering
  \centerline{\includegraphics[width=12cm]{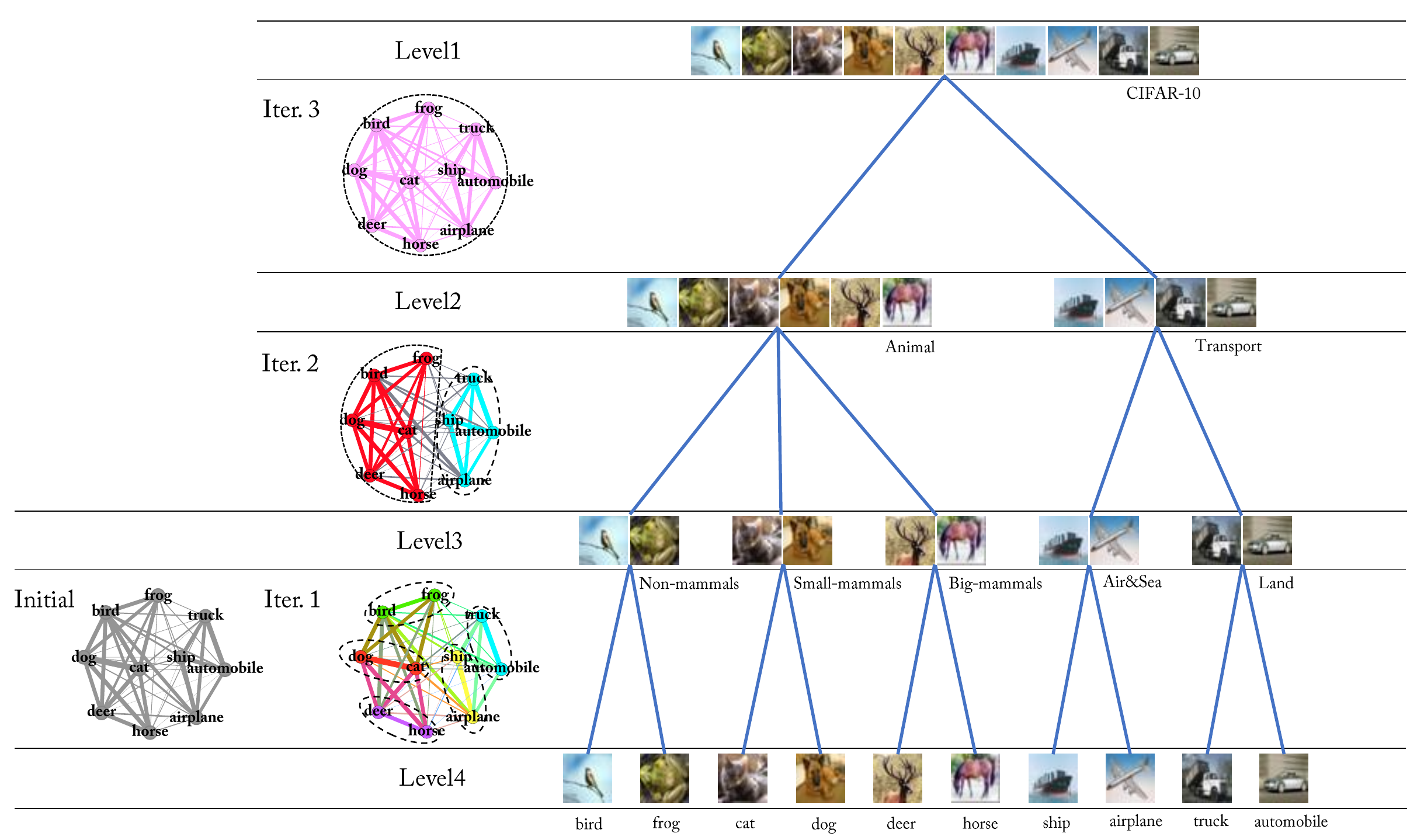}}
\end{minipage}
\caption{The construction process of confusion visual tree for CIFAR-10 image set.}
\label{fig:figure2}
\end{figure*}

CNN models have been successfully applied in many computer vision tasks such as image classification. Recently, there are some works focusing on linking the structure of CNN models with the structure of tree models. One of the earliest attempts is reported in~\cite{Srivastava2013Discriminative} but their main goal is transferring knowledge between categories to improve the results for categories with insufficient training examples. In~\cite{Yan2015HD}, the proposed model HD-CNN embeds a category hierarchy into CNN models so the model separates easy categories using a coarse-grained category classifier while distinguishing difficult categories using fine-grained category classifiers. The Branch Convolutional Neural Network(B-CNN) model, proposed in~\cite{Zhu2017B}, outputs multiple predictions ordered from coarse-grained to fine-grained along the concatenated convolutional layers corresponding to the hierarchical structure of the target categories, which can be regarded as a form of prior knowledge on the output. However, the HD-CNN model has to pretrain the shared layers and fine-tune the pre-trained models on fine-grained category components several times, which is much complicated and there is no close connection between the coarse-grained category branches and the fine-grained category branch in the B-CNN model except for their loss weights.

Our VT-CNN model has a more simple network structure compared with the HD-CNN model so we can train VT-CNN without fine-tuning the pre-trained models and we use a new loss function in the fine-grained category branch in order to connect all the branches closely.

\section{Method}
\label{sec:m}
We explain the overview of our new image classification method called VT-CNN. First we introduce the method to construct CVT. Then we describe the details of VT-CNN.

\subsection{Establish a Confusion Visual Tree}\label{sec:cvt}
The goal of this stage is to construct the visual tree for the VT-CNN model. The tree structure in our VT-CNN model, which we call it Confusion Visual Tree, is established in three steps as follows. The first is using the confusion graph generation algorithm~\cite{Jin2017Confusion} to compute confusion graph and its weights. The second is using the community hierarchical detection algorithm~\cite{Vincent2008Fast} to generate communities in the confusion graph. The last is building a CVT by utilizing the results of the second step.

Here we talk about the details in the construction of the CVT. We apply the confusion graph generation algorithm on the dataset $D$ for a CNN model $M$ and get the confusion graph $G$. Then we apply the community hierarchical detection algorithm on $G$ and we get the community set $C$. Notice that the $C$ has two dimensions. The one refers to the index of the outputs of all the iterations of the algorithm and the other refers to the communities in each output. As for the tree $T=(V,E,L)$, $V$ denotes the nodes in the tree and leaves refer to the categories and other nodes refer to the symbol of the communities, $E$ denotes the edges that connecting the nodes in a high level in $C$ to nodes in a lower level in $C$ while the former nodes belong to the community sets represented by the latter nodes and $L$ is the label set in each node which can be generated by the nodes' corresponding community set.

We expound the CVT construction process specifically on \emph{CIFAR-10} dataset which is shown in Fig. \ref{fig:figure2}. The confusion graph and the communities inside are shown on the left side and the CVT on the right side. The left side is divided into four steps: Initial and Iter. $1$ to $3$. We use the confusion graph generation function to generate a confusion graph which is shown at the Initial step. Each vertex represents one category in the dataset and the weight of each edge quantifies the confusion between two connected categories. Then we apply the community hierarchical detection function on the confusion graph and we get a community graph at each iteration of this function. At Iter. $1$ step, for instance, we get five fine-grained communities and we set five nodes to the tree in "level$3$" which correspond to the communities one by one. Each member of one of the communities refers to a specific category. Then we link the leaves to level$3$ nodes. For instance, we link node "cat" and node "dog" to node "Small-mammals" in level$3$. We repeat it until nodes in level$2$ are linked to root node in level$1$ and then we finish the construction process.

\subsection{VT-CNN Architecture}
In this section, we embed the CVT into an original CNN model to construct the VT-CNN model. Take the classification task on CIFAR-$100$ as an example, the architecture of the VT-CNN model is shown in Fig. \ref{fig:figure1a} and the $4$ levels CVT of CIFAR-$100$ is in Fig. \ref{fig:figure1b}. The goal of the VT-CNN model is to distinguish the fine-grained categories presented as leaves of the tree structure, which are the final targets of the classification task. For the tree structure, we use \emph{level-$n$} to refer to the $n$th layer of the visual tree. We define that level-$1$ refers to the root node and level-$N$ refers to the leaves of the tree which has $N$ layers.

The VT-CNN model is based on an existing CNN model. It has two important components. The one is the trunk layers, which we called the \emph{base} architecture. The other is the branch layers and we name them \emph{branch} architecture. The \emph{base} architecture is actually borrowed from an original CNN model such as the AlexNet model and the VGG-Verydeep-16(VGG16) model and the layers in it are shared by all of the branches. The \emph{branch} architecture is related to different levels of the visual tree. The architecture in each branch contains two ConvNets and two FC layers and each branch is associated with a discrimination task for its related level in the tree architecture. The number of the layers in the \emph{branch} architecture is equal to the number of the levels in the visual tree without the root level. Followed~\cite{Zhu2017B}, low layers in CNN models usually capture the low level features of an image such as basic shapes while higher layers are likely to extract higher level features such as the face of the dog. Similar to this conception, the coarse-grained categories on low level of the tree architecture is associated with the low level features mentioned above while the fine-grained categories on leaves with the high level features. Therefore, the branch layers which are extended from low layers in CNN models are associated with the low level in the tree architecture while high layers with high level in the tree architecture.

The tree structure of the VT-CNN model provides the prior information contained in branches of low layers for fine-grained categories classification in the last branch of high layers. It is benefit from the conception that the information on the leaves of a tree structure which is the key to the fine-grained classification is actually contained in their ancestors which is responsible for the coarse-grained classification. Therefore, our VT-CNN focuses on the fine-grained categories in same communities instead of treating all the categories equally, which is different to original CNN models.

\subsection{VT-CNN Training}
The idea of our VT-CNN model is that the prior information of the coarse-grained categories classification should be used for the fine-grained categories classification. So we split the training of our VT-CNN into three steps which are the establishing of the CVT, the training of coarse-grained branches and the training of the fine-grained branch.

\subsubsection{Establish a Confusion Visual Tree for a Specific CNN model}\label{sec:traincvt}
We firstly establish a CVT for a CNN model which is used as the basic CNN model of our VT-CNN model. Firstly, we train the original CNN model on a specific dataset and we get its outputs which are scores on each category of images in the dataset. Then we use the algorithm from Section \ref{sec:cvt} to establish the CVT based on these outputs.

\subsubsection{Training Data}
In order to train the VT-CNN model, there should be some changes in the original dataset. Each image should have several labels which are consistent with the coarse-grained and fine-grained category divisions based on the CVT constructed in Section \ref{sec:traincvt}. The number of the labels is equal to the number of branches in the VT-CNN model. Take an image whose label is ''dog'' in CIFAR-$10$ as an example, the image has three labels which are ''Animal'', ''Small-mammals'', ''Dog'' according to the rule mentioned above.

\subsubsection{Train the Coarse-grained Branches}
All the branches except for the last branch which is associated with the fine-grained categories classification are training in this step. We give the loss function directly. The loss function in the training of the VT-CNN model contains all the loss function in coarse-grained branches. The loss function is defined as:
\begin{equation}
L_C = -\frac{1}{n}\sum^n_{i=1}\sum^{K-1}_{k=1}{W_k\log\left(\frac{e^{c^k_{y_i}}}{\sum_j{e^{c^k_j}}}\right)}
\end{equation}
where $i$ denotes the $i^{th}$ sample in the mini-batch, $K$ is the number of all the branches in the VT-CNN model, $W_k$ is the loss weight of $k$th branch contributing to the loss function and $c_j$ denotes the $j$th element in the vector $c$ of class scores. Notice that we have no need to consider the last branch in this section so the number of the branches is from $k=1$ to $k=K-1$. This loss function $L_C$ computes the softmax cross-entropy on these branches and add them together.

\subsubsection{Train the Fine-grained Branch}
After the coarse-grained branches are properly trained, we start training the fine-grained branch which is also the last branch of the VT-CNN model. In the probabilistic averaging layer, we change the final prediction into a weighted average prediction which is computed from the $(K-1)$th coarse-grained branch as below:
\begin{equation}
f_i = \frac{c^{K-1}_{t(i)}f^K_i}{\sum_j{c^{K-1}_{t(j)}f^K_j}}
\end{equation}
where $t$ is the fine-grained category to the $(K-1)$th coarse-grained category set so $t(i)$ denotes the related coarse-grained category of the fine-grained category $i$.

The loss function of the fine-grained branch is defined as.
\begin{equation}
L_F = -\frac{1}{n}\sum^n_{i=1}{W_K\log\left(\frac{e^{f_{y_i}}}{\sum_j{e^{f_j}}}\right)}
\end{equation}

During the training process, we change the weight $W_k$ to control that which branch is going to be trained. In our training strategy, we want to train the coarse-grained branches firstly so we firstly set $\sum^{K-1}_{k=1}{W_k} = 1$ and $W_K = 0$. Then we train the fine-grained branch and we set $\sum^{K-1}_{k=1}{W_k} = 0, W_K = 1$. Finally, we finish the training process.

\section{Experiment}
\label{sec:exp}
\subsection{Datasets and Experiment Settings}
To comprehensively evaluate our proposed method, the following two image datasets are used in our experiment: CIFAR-$10$ and CIFAR-$100$.

The CIFAR-$10$ is a dataset for general object recognition which has $10$ categories. Each image is natural RGB with $32\times 32$ pixels and the dataset has $60000$ images in total. $50000$ for training and $10000$ for testing. The CIFAR-$100$ dataset has $60000$ images and $100$ categories in total. Each image is natural RGB with $32\times 32$ pixels and each category has $600$ images in which $500$ for training and $100$ for evaluating.  Our experiment on these two datasets follows their division.

The Top-$1$ Mean Accuracy $(\%)$ is used as the criterion to evaluate the performance of all these approaches. A PC with Intel Core i7, 32GB memory and NVIDIA GeForce GTX $1080$ Ti is utilized to run all the experiments.

\subsection{CIFAR-10}
\label{sec:exp2}

Experiments on CIFAR-$10$ dataset compare the VT-CNN models with base models and the B-CNN~\cite{Zhu2017B} models.

We construct two baseline models. The Base A model is actually the AlexNet~\cite{Krizhevsky2012ImageNet} model and there is just a little different between them that the size of filters in the first and second convolutional layers of the Base A model is adapted to $5\times 5$. The Base B model is the VGG$16$~\cite{Simonyan2014Very} model without the final max-pooling layer because images in the CIFAR dataset are very small.

Then we construct VT-CNN models based on the baseline models. In CIFAR-10 dataset, our CVT model has $4$ hierarchical layers so there is $3$ branches in VT-CNN models. The specific configuration is shown in Table \ref{tab:exp1a}.

\begin{table}[t]
	\captionsetup{format=plain,labelsep=colon,justification=raggedright}
    \centering
    \setlength{\abovecaptionskip}{0pt}
	\setlength{\belowcaptionskip}{10pt}
    \caption{The entire architecture for each VT-CNN model. The id number in symbol conv-$id$ denotes this part of branch architecture is connected with the $id$th convolution layers. Note that the branch layers are illustrated in bold fonts.}
    \normalsize
    \begin{tabular}{|c|c|}
        \hline
        Layers of Base A&Layers of Base B\\
        \hline
        \hline
        (Conv3-98)-MaxPool&(Conv3-64)$_{\times 2}$-MaxPool\\
        \hline
        (Conv3-256)-MaxPool&(Conv3-128)$_{\times 2}$-MaxPool\\
        \hline
        (Conv3-384)&(Conv3-256)$_{\times 3}$-MaxPool\\
        \hline
        \multicolumn{2}{|c|}{\bf{conv-$3$ Flatten}}\\
        \hline
        \bf{(Conv3-128)$_{\times 2}$}&\bf{(Conv3-256)$_{\times 2}$}\\
        \hline
        \bf{FC-256}&\bf{FC-512}\\
        \hline
        \bf{FC-256}&\bf{FC-512}\\
        \hline
        \bf{FC-$C_{A1}$}&\bf{FC-$C_{B1}$}\\
        \hline
        (Conv3-384)&(Conv3-512)$_{\times 3}$-MaxPool\\
        \hline
        \multicolumn{2}{|c|}{\bf{conv-$4$ Flatten}}\\
        \hline
        \bf{(Conv3-256)$_{\times 2}$}&\bf{(Conv3-512)$_{\times 2}$}\\
        \hline
        \bf{FC-512}&\bf{FC-1024}\\
        \hline
        \bf{FC-512}&\bf{FC-1024}\\
        \hline
        \bf{FC-$C_{A2}$}&\bf{FC-$C_{B2}$}\\
        \hline
        (Conv3-256)-MaxPool&(Conv3-512)$_{\times 3}$-MaxPool\\
        \hline
        \multicolumn{2}{|c|}{conv-$5$ Flatten}\\
        \hline
        \multicolumn{2}{|c|}{Probabilistic averaging layer}\\
        \hline
    \end{tabular}
    \label{tab:exp1a}
\end{table}

We evaluate our VT-CNN model, their base models and B-CNN models on CIFAR-10 dataset. Notice that the tree structure of the B-CNN model is the same as that in our VT-CNN model. For model A, both models' learning rates are initialized to be $0.003$ and decease to $0.0005$ after $40$ epochs and $0.0001$ after $50$ epochs. For model B, we fine-tune on the pre-trained VGG$16$ model on the ImageNet dataset and the configuration of learning rates is the same as model A. For loss weights in the coarse-grained branches training step, we set them all the same.

\begin{table}[h]
	\captionsetup{format=plain,labelsep=colon,justification=raggedright}
    \centering
    \setlength{\abovecaptionskip}{0pt}
	\setlength{\belowcaptionskip}{10pt}
    \caption{Performance of each model on CIFAR-10 test set.}
    \normalsize
    \begin{tabular}{|l|c|}
        \hline
        Models&Top-$1$ Accuracy\\
        \hline
        Base A&$82.94\%$\\
        \hline
        B-CNN A&$84.70\%$\\
        \hline
        \textbf{VT-CNN A}&$\mathbf{85.07\%}$\\
        \hline
        \hline
        Base B&$87.15\%$\\
        \hline
        B-CNN B&$88.23\%$\\
        \hline
        \textbf{VT-CNN B}&$\mathbf{89.51\%}$\\
        \hline
    \end{tabular}
    \label{tab:exp1b}	
\end{table}

The results are shown in Table \ref{tab:exp1b}. The baseline model A gets an accuracy of $82.94\%$ and the B-CNN model achieves $84.70\%$. Our VT-CNN model beats the B-CNN model by an accuracy of $85.07\%$. The accuracy in Base B model is $87.15\%$ while the B-CNN model reaches $88.23\%$. Our VT-CNN model achieves $89.51\%$. The improvement is obvious which indicates that the performance of our VT-CNN models is better than original CNN models and the B-CNN models on CIFAR-$10$ dataset.

\subsection{CIFAR-100}
In this section, we evaluate our VT-CNN model on CIFAR-100 dataset. The experiment is consist of two parts. The one is the comparison with VT-CNN models, baseline models of $3$ original CNN models and B-CNN models based on these baseline models. The other is the comparison among VT-CNN models constructed on different visual tree structures.

First, we introduce $3$ original CNN models which are AlexNet, VGG$16$ and ResNet-$56$. The configurations for the AlexNet model and the VGG$16$ model on CIFAR-$100$ dataset are the same as the Base A model and the Base B model in Section \ref{sec:exp2}. The ResNet-$56$~\cite{He2015Deep} is a residual learning network of depth $56$ using residual blocks of $2$ convolutional layers.

We construct VT-CNN models based on these baseline models. Their branch architecture configurations are similar with the configuration in Section \ref{sec:exp2} because the main idea of the VT-CNN model is that the branches should connect to different positions on the base layers. There is a little different that our CVT on CIFAR-$100$ dataset has $5$ layers so there are $4$ branches in VT-CNN models.

\begin{table}[h]
	\captionsetup{format=plain,labelsep=colon,justification=raggedright}
    \centering
    \setlength{\abovecaptionskip}{0pt}
	\setlength{\belowcaptionskip}{10pt}
    \caption{Performance of each model on CIFAR-100 test set.}
    \normalsize
    \begin{tabular}{|l|c|}
        \hline
        Models&Top-$1$ Accuracy\\
        \hline
        AlexNet A&$57.37\%$\\
        \hline
        B-CNN A&$58.27\%$\\
        \hline
        \textbf{VT-CNN A}&$\mathbf{58.73\%}$\\
        \hline
        \hline
        VGG$16$ B&$71.15\%$\\
        \hline
        B-CNN B&$71.23\%$\\
        \hline
        \textbf{VT-CNN B}&$\mathbf{72.04\%}$\\
        \hline
        \hline
        ResNet-$56$ C&$73.49\%$\\
        \hline
        B-CNN C&$73.86\%$\\
        \hline
        \textbf{VT-CNN C}&$\mathbf{74.13\%}$\\
        \hline
    \end{tabular}
    \label{tab:exp2b}	
\end{table}

These $3$ baseline models and their derived B-CNN and VT-CNN models are trained in $80$ epochs and using the different learning rates at different epochs. The learning rates are initialized to be $0.001$ and decease to $0.0002$ after $55$ epochs and $0.00005$ after $70$ epochs. The complete experiment results are given in Table \ref{tab:exp2b}. In every case, our VT-CNN model achieves highest accuracy than the corresponding base model and the derived B-CNN model. In the case of AlexNet, the relative improvement is $1.36\%$ compared with the base model and $0.46\%$ compared with the B-CNN model . The VGG$16$ model is used as base model to train the VT-CNN model. The VGG$16$ model gives $71.15\%$ and the derived B-CNN model gives $71.23\%$,  whereas we achieve $72.04\%$. Finally, we consider about the state-of-the-art technique ResNet as the baseline model and use the ResNet-$56$ model. The accuracy of $74.13\%$ obtained by our VT-CNN model built from the ResNet-$56$ model is the best result in this experiment while the ResNet-$56$ model reaches $73.49\%$ and the B-CNN model achieves $73.86\%$. We find that our VT-CNN model achieves the best performance in the classification task on the CIFAR-$100$ dataset compared with the original CNN models and their derived B-CNN models.

Then we do experiment on comparison of VT-CNN models constructed on different visual tree structures. The following competing tree structures are chosen for comparison: {\bf Semantic Ontology~\cite{Li2010Building}, Label Tree~\cite{Bengio2010Label}, Visual Tree~\cite{Fan2015Hierarchical}} and the state-of-the-art technique {\bf Enhanced Visual Tree~\cite{Zheng2017Hierarchical}}. The classification performance of all the VT-CNN based on these structures are reported in Table \ref{tab:exp2c}.

\begin{table}[h]
	\captionsetup{format=plain,labelsep=colon,justification=raggedright}
    \centering
    \setlength{\abovecaptionskip}{0pt}
	\setlength{\belowcaptionskip}{10pt}
    \caption{Results on VT-CNN with different tree structures.}
    \normalsize
    \begin{tabular}{|l|c|}
        \hline
        Tree Structure in VT-CNN&Accuracy\\
        \hline
        Semantic Ontology&$57.18\%$\\
        \hline
        Label Tree&$64.39\%$\\
        \hline
        Visual Tree&$69.48\%$\\
        \hline
        Enhanced Visual Tree&$72.14\%$\\
        \hline
        \textbf{Confusion Visual Tree}&$\mathbf{74.13\%}$\\
        \hline
    \end{tabular}
    \label{tab:exp2c}	
\end{table}

Observing Table \ref{tab:exp2c}, we find the performance of Semantic Ontology is the worst because its tree structure is constructed based on semantic space and the image classification process based on feature space~\cite{Fan2015Hierarchical}. For another four methods based on feature space, the performance of the Label Tree is worse because of using OvR classifier to construct its tree structure, which is limited to sample imbalance and the performance of classifier. As for the Visual Tree, it uses the average features extracted directly from the dataset. The Enhanced Visual Tree adopts the spectral clustering method that better reflects the diversity of categories, so its performance is better than the Visual Tree~\cite{Zheng2017Hierarchical}. Our CVT constructs the tree structure based on the confusion of the neural network, which makes sibling nodes as close as possible to the parent node as far as possible. So the structure of the CVT is more proper and we obtain a significant improvement over the Visual Tree and the Enhanced Visual Tree by $4.65\%$ and $1.99\%$.

\section{Conclusion}
In this paper, we propose a Visual Tree Convolution Neural Network(VT-CNN) which connects the original CNN models with the visual tree. Compared with original CNN models, VT-CNN models can utilize the prior information in coarse-grained category classification to improve the performance of the final fine-grained category classification. We also introduce a method to construct a Confusion Visual Tree(CVT) which provides a more reasonable tree structure for the VT-CNN models. The experiment results confirm the benefits of our VT-CNN model with CVT over the original CNN model. For further work the improvement on the final fine-grained classification layers, such as splitting the all-category FC-softmax layer to several fine-grained category sets FC-softmax layers, should be investigated.

\bibliographystyle{IEEEtran}
\bibliography{IEEEexample}

\end{document}